\documentclass[journal]{IEEEtran}

\usepackage{xcolor,framed} 
\usepackage{color}
\usepackage{soul}
\usepackage{bbding} 
\usepackage{multirow}
\usepackage{booktabs}

\colorlet{shadecolor}{yellow}
\usepackage[pdftex]{graphicx}
\graphicspath{{../pdf/}{../jpeg/}}
\DeclareGraphicsExtensions{.pdf,.jpeg,.png}

\usepackage[cmex10]{amsmath}
\usepackage{array}
\usepackage{mdwmath}
\usepackage{mdwtab}
\usepackage{makecell}
\usepackage{eqparbox}
\usepackage{pifont}
\usepackage{url}
\usepackage{graphicx}
\usepackage{amsmath}
\usepackage[caption=false,font=footnotesize,labelfont=rm,textfont=rm]{subfig}
\usepackage{caption}
\usepackage{float}
\usepackage{floatflt}
\usepackage{amssymb}
\usepackage[
pdfauthor={derajan},
pdftitle={How to do this},
pdfstartview=XYZ,
bookmarks=true,
colorlinks=true,
linkcolor=blue,
urlcolor=blue,
citecolor=blue,
pdftex,
bookmarks=true,
linktocpage=true,   
hyperindex=true
]{hyperref}

\begin{document}
\bstctlcite{IEEEexample:BSTcontrol}
    \title{Enhancing, Refining, and Fusing: Towards Robust Multi-Scale and Dense Ship Detection}
  \author{Congxia Zhao,
          Xiongjun Fu,~\IEEEmembership{Senior Member,~IEEE,}
          Jian Dong,
          Shen Cao,
          and Chunyan Zhang

  \thanks{ 
  \textit{(Corresponding authors: Xiongjun Fu and Jian Dong.)}}%
  \thanks{Congxia Zhao, Xiongjun Fu,  Jian Dong, Shen Cao, and Chunyan Zhang are with the School of Integrated Circuits and Electronics, Beijing Institute of Technology, Beijing (e-mail: zcx09890989@163.com; fuxiongjun@bit.edu.cn; radarvincent@sina.com; caoshen@iscas.ac.cn; 312023133@bit.edu.cn).}}%



\markboth{
}{Roberg \MakeLowercase{\textit{et al.}}: High-Efficiency Diode and Transistor Rectifiers}

\maketitle

\begin{abstract}
Synthetic aperture radar (SAR) imaging, celebrated for its high resolution, all-weather capability, and day-night operability, is indispensable for maritime applications. However, ship detection in SAR imagery faces significant challenges, including complex backgrounds, densely arranged targets, and large scale variations. To address these issues, we propose a novel framework, Center-Aware SAR Ship Detector (CASS-Det), designed for robust multi-scale and densely packed ship detection.
CASS-Det integrates three key innovations: (1) a center enhancement module (CEM) that employs rotational convolution to emphasize ship centers, improving localization while suppressing background interference; (2) a neighbor attention module (NAM) that leverages cross-layer dependencies to refine ship boundaries in densely populated scenes; and (3) a cross-connected feature pyramid network (CC-FPN) that enhances multi-scale feature fusion by integrating shallow and deep features. 
Extensive experiments on the SSDD, HRSID, and LS-SSDD-v1.0 datasets demonstrate the state-of-the-art performance of CASS-Det, excelling at detecting multi-scale and densely arranged ships. 
\end{abstract}


%
\IEEEpeerreviewmaketitle


\section{Introduction}

Synthetic aperture radar (SAR) is widely utilized in both military and civilian applications due to its capability to generate high-resolution microwave images irrespective of time and weather conditions \cite{1, 2, 3}. It plays a crucial role in national security, earth observation, and environmental monitoring. In the maritime domain, SAR ship detection has become essential for applications such as maritime traffic control, rescue operations, environmental protection, and other related tasks \cite{4, 5, 6}.

Traditional SAR ship detection methods involve several sequential steps: preprocessing, sea-land segmentation, candidate region extraction, and target recognition. Preprocessing and sea-land segmentation aim to mitigate speckle noise and reduce the influence of land clutter. Features are then extracted in spatial and transform domains \cite{10, 11, 12, 13, 14}, and potential targets are identified using constant false alarm rate (CFAR) algorithms. CFAR adjusts local thresholds based on background clutter and compares pixel intensities against these thresholds to detect targets. However, its reliance on accurate statistical modeling of background clutter makes it highly sensitive to noise interference and unsuitable for complex scenes, often resulting in unstable detection performance \cite{15}.

\begin{figure}[t]
\setlength{\belowcaptionskip}{-0.3cm}
\begin{center}
  \includegraphics[width=0.95\linewidth]{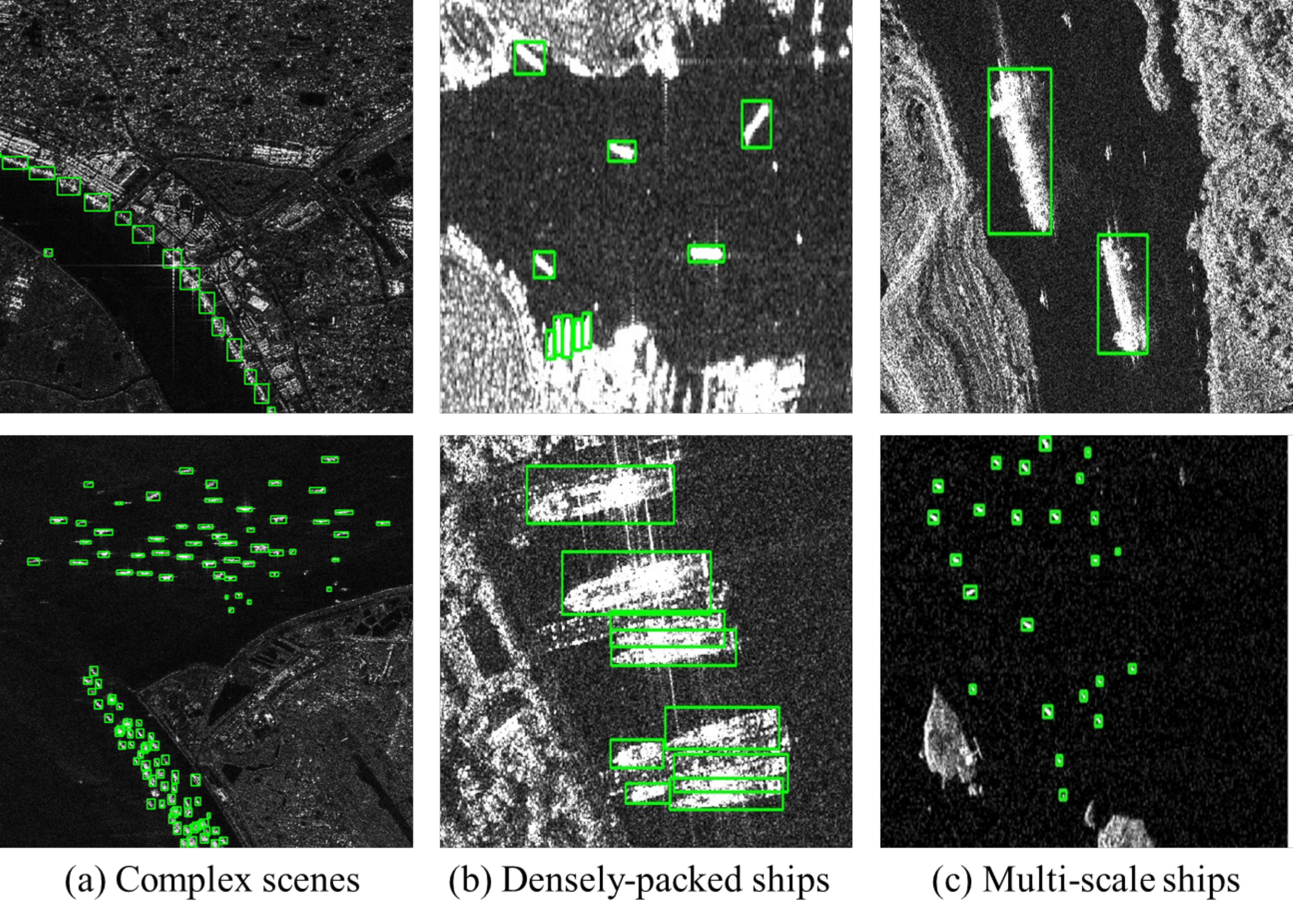}
\captionsetup{font={footnotesize}}
\caption{Images from SSDD and HRSID. The ground truth in the images are marked with green bounding boxes.}
\label{exa_from_SSDD}
\end{center}
\end{figure}

With recent advancements in deep learning, convolutional neural networks (CNNs) \cite{19} have shown remarkable success due to their powerful feature extraction capabilities. CNN-based object detection methods are widely adopted for their high efficiency, accuracy, and robustness. These methods generally fall into two categories: two-stage and one-stage detectors. Two-stage methods first generate region proposals and then classify and locate targets within these proposals. Examples include Faster R-CNN \cite{20}, SPP-Net \cite{21}, and Cascade R-CNN \cite{22}. In contrast, one-stage detectors directly predict target locations and categories from extracted features, enabling real-time processing. Popular one-stage detectors include SSD \cite{23}, YOLO series \cite{24, 25, 26, 27, 28}, RetinaNet \cite{29}, FCOS \cite{30}, and CenterNet \cite{31}. Both approaches continue to evolve towards achieving higher precision and faster inference speeds.

CNN-based methods have become prevalent in SAR ship detection \cite{SSE, Light, 39}, outperforming traditional approaches. However, challenges persist in this domain:
(1) \textbf{Complex Scenes}: SAR imagery often contains intricate backgrounds, such as coastlines, ports, and rivers, complicating the delineation of ship boundaries and causing confusion with similar-shaped objects, leading to missed detections.
(2) \textbf{Densely Arranged Inshore Ships}: Inshore ships often appear densely packed, touching or overlapping one another, which causes detectors to mistake adjacent ships for a single target, reducing recall and precision.
(3) \textbf{Large-Scale Variations}: Ship sizes vary significantly due to differences in their lengths and imaging resolutions, resulting in challenges when detecting targets of various scales and orientations.

Previous approaches have addressed these issues through attention mechanisms and multi-level feature fusion. However, standalone attention modules that focus solely on ship areas struggle to separate densely packed ships. Similarly, conventional feature fusion modules, such as feature pyramid network (FPN) \cite{41} and path aggregation feature pyramid Network (PAFPN) \cite{42}, typically consist of three layers and fail to accommodate the wide range of ship sizes. Increasing the number of fusion layers indiscriminately leads to a substantial rise in computational costs.
To overcome these challenges, we propose CASS-Det, a robust one-stage SAR ship detector for densely packed and multi-scale targets. The main contributions of this work are as follows:
\begin{itemize}
    \item We design a Center Enhancement Module (CEM) based on rotational convolution to emphasize ship centers. By extracting and stacking features from multiple directions, CEM highlights central features and suppresses background interference, effectively guiding the network's attention to target regions.
    \item We introduce a Neighbor Attention Module (NAM) to refine boundaries between densely packed ships. NAM leverages long-range relationships of different levels to combine different fine-grained global features, enabling accurate boundary distinction and improving recall.
    \item We propose a Cross-Connected Feature Pyramid Network (CC-FPN) to address large-scale variations in ship sizes. The CC-FPN extends feature fusion layers using a cross-connected structure, integrating shallow detail information with deep semantic features while minimizing computational overhead.
    \item We validate CASS-Det on the SSDD \cite{43}, HRSID \cite{44}, and LS-SSDD-v1.0 \cite{LS-SSDD} datasets. Experimental results demonstrate that CASS-Det significantly outperforms state-of-the-art algorithms in detecting multi-scale and densely packed ships in SAR images.
\end{itemize}


\section{Related Work}

\subsection{CNN-based SAR Ship Detection}

Traditional SAR ship detection methods are mainly based on the feature extraction and target recognition. Although these methods perform well in specific conditions, they do not adapt to all complex environments. Moreover, CNN-based methods can independently extract effective features from images, with great universality and stability. Therefore, in recent years, SAR object detection algorithms based on CNNs have attracted many researchers' attention and study.

In SAR ship detection, researchers adjust the structure of models to make them more suitable for the characteristics of the ships and backgrounds. 
To deal with the noise from inshore and inland interferences, a spatial shuffle-group enhance (SSE) attention module \cite{SSE} is developed and put into the network based on the CenterNet. 
A transformer-based dynamic sparse attention module, small target-friendly detection heads, and loss function are proposed to contrusted a lightweight network\cite{Light}. Typical designing scheme improves the detection effect of small targets. 
For the fuzzy and complex surroundings in the SAR image, non-subsampling Laplacian pyramid decomposition (NSLP) is utilized  as a pre-processing step to effectively extract detailed features. These features were then seamlessly integrated into CNNs, yielding outstanding results. 
SRT-Net \cite{40} introduced the gragh convolutional networks (GCNs) into the CNNs to extract the ships' structural features. GCNs provide feature similarity between regions to help distinguish ships from other objects, thereby reducing false detection and improving detection accuracy.

Above methods improve the suitability of the intelligent networks in SAR ship detection and improve the detection effect. However, multi-scale ships and complex background in the SAR images bring obstacles to detection.

\subsection{Multi-scale Targets Detection}

There are many kinds of ships in the ocean, including fishing boats, freighters, sundries, passenger ships, warships and so on. These ships vary greatly in size, from 2 meters to several hundred meters. In addition, the detectors for imaging have various resolution, resulting in a huge difference in the ships' sizes in SAR images. The smallest ship occupies only a few pixels, while the large ship covers hundreds of pixel values. Different sizes bring great difficulties to the accurate detection. 

In the natural object detection, FPN and PAFPN are typically designed for multi-scale object detection. FPN builds a top-down branch to fuse different feature maps from different levels and transmit deep semantic information to shallow layers. On the contrary, PAFPN provides a down-top branch transmitting deep semantic information to shallow layers further. However, common three-layer structure cannot meet the gap of ships' scales in SAR images. 
Some researchers solve this obstacle through improved feature fusion structure. 
Dense attention pyramid network (DAPN) \cite{33} enhanced multi-scale ship detection in complex SAR imagery by integrating convolutional block attention modules (CBAM) \cite{34} with cascading feature maps of a pyramid network, thereby capturing a rich set of features that include both fine details and semantic information. 
\cite{38} proposed a receptive field increased module to extract multi-scale feature maps and construct a spatial pyramid enriched with scale information, facilitating the detection of ships across various scales and minimizing missed detections.
Some researchers develop attention modules to obtain information for multi-scale targets. 
Balance attention network (BANet) \cite{BANET} introduced local and non-local attention models to acquire detailed and semantic information.   
The improvements of detection results from the above methods indicate that both ways can improve the detection effect of multi-scale ships.

\begin{figure*}[t]
\setlength{\belowcaptionskip}{-0.5cm}
  \begin{center}
  \includegraphics[width=0.95\linewidth]{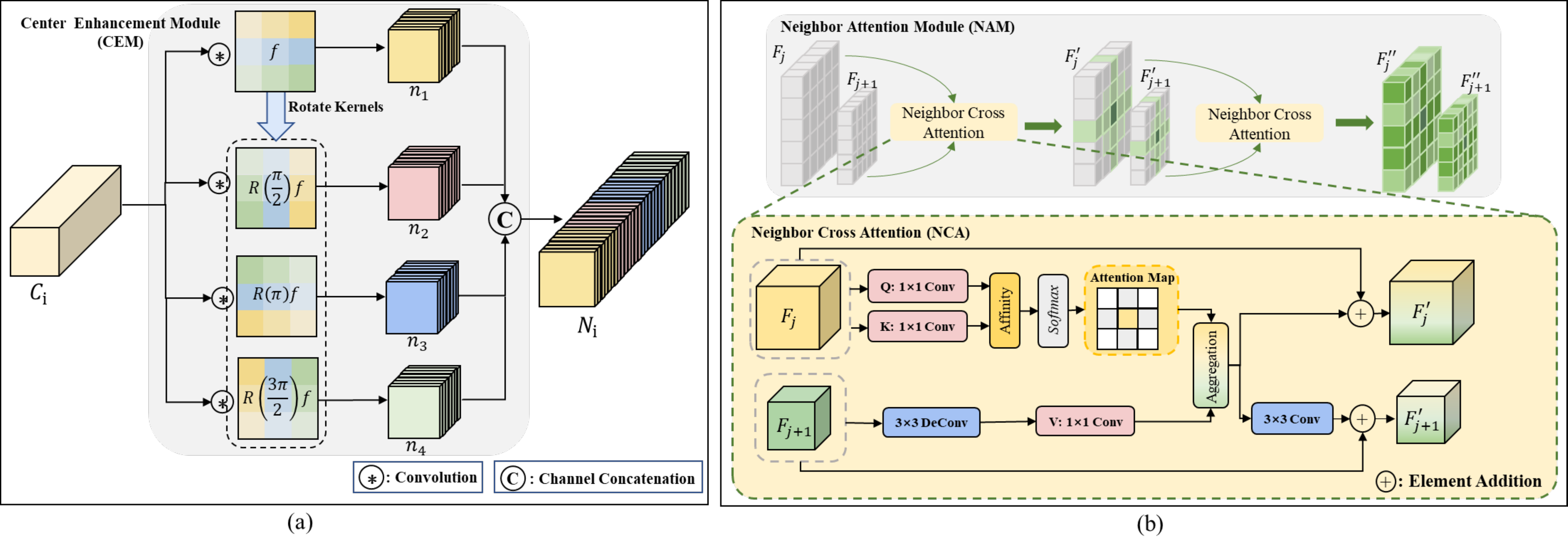}
  \captionsetup{font={footnotesize}}
  \caption{The structure of center enhancement module (CEM) and neighbor attention module (NAM). In (a), \(C_i\)(\(i = 2, 3, 4, 5\)) presents the input. \(f\) is the convolution kernel. \(R(k)f\) represents the convolution kernel \((f)\) rotated with \(k\). \(n_1, n_2, n_3, n_4\) are the convolution results of \(C_i\) and rotated kernels. In (b), long-range dependencies of the image are calculated based on the feature maps of two adjacent layers, and obtain the global features of different fine-grained synthesis. \(F_j\) and \(F_{j+1}\) are inputs of NAM. \(F'_j\) and \(F'_{j+1}\) reflect the the relationship between each pixel and the pixels in horizontal and vertical directions. With the recurrent neighbor cross attention module (NCA), \(F''_j\) and \(F''_{j+1}\) contain the global information.}
  \label{sub}
  \end{center}
\end{figure*}

\subsection{Complex Backgrounds in SAR Images}

Ships have different motion states, either traveling or docked, and appear in various SAR images over oceans, rivers, and harbors\cite{35,36,37,38}. Speckle noise from the SAR principle, along with similar-looking islands, can lead to false detections. Additionally, coastlines often share similar intensity levels with ships, causing false positives or missed detections. Diffusion models \cite{shen2024imagpose, shen2024imagdressing} can help mitigate these issues by refining detection results and reducing noise, enhancing accuracy in complex backgrounds.
The researchers\cite{shen2024boosting,36, shen2023advancing} use spatial domain analysis, transform domain analysis, and deep learning methods to remove speckle noise. Moreover, long-range dependencies are helpful to deal with the confusing marine and land environment in the image. Many researchers have achieved fruitful results in this field. 
\cite{Non-local} opened the era of extracting global information with non-local neural network and applied it to image processing. 
The global context (GC) block \cite{GCNet} optimized the overall network architecture by simplifying the process of feature map acquisition and reduced the amount of network computation. 
Dual attention network (DANet) \cite{DAnet} combined spacial and channel attention with non-local, fully exploring the importance of spatial attention and channel attention. 
The criss-cross attention module \cite{CCNet} adopted the contextual information of the pixels on the criss-cross path, achieving a similar effect of non-local with reduced computation. These methods can synthesize the global information, distinguish the boundary between the surroundings and the targets, and accurately determine the location of the targets.

\section{Methodology}

This section introduces the proposed method, which consists of three primary components: the Center Enhancement Module (CEM), the Neighbor Attention Module (NAM), and the Cross-Connected Feature Pyramid Network (CC-FPN). These components address the challenges of background clutter, densely packed targets, and scale variations in SAR ship detection. The following subsections provide detailed descriptions of each module and their contributions.

\subsection{Center Enhancement Module}
SAR imagery often contains complex backgrounds, including land clutter, sea clutter, and speckle noise, which obscure ship features and degrade detection accuracy. While these interferences exhibit irregular textures and intensity variations, ships generally display distinct edges, structural integrity, and higher grayscale values. These characteristics enable ships to stand out in SAR imagery compared to clutter and noise, which lack defined contours and central focus. The Center Enhancement Module (CEM) leverages these properties by applying rotational convolution to enhance ship centers and suppress background clutter.

{(1) \textit{Rotational Convolution}: Traditional 2D convolution lacks rotational invariance, as signal rotation involves interpolation and sampling. The convolution of a rotated feature map is equivalent to the standard convolution with an inversely rotated kernel:
\begin{equation}
    [a * [\text{R}b]](x) = \text{R}^{-1}[(\text{R}a) * b](x).
\end{equation}
Where $a, b$ represent feature map and convolution kernel. $\text{R}(\cdot)$ denotes rotation, and $*$ represents convolution.

(2) \textit{Structure of CEM}: As shown in Fig.~\ref{sub} (a), the CEM focuses on ships' centers through rotational convolution with rotated kernels and concatenation. Based on the derivation in (1), the module achieves rotational convolution in feature extraction by rotating the input feature map $C_{i} \in \mathbb{R}^{C \times H \times W}$, where $C$ is the number of channels, and $H$ and $W$ are the height and width of the feature map. The rotated feature maps are convolved with the kernel $f$ and concatenated across four orientations:
\begin{equation}
\begin{aligned}
N_{\text{i}} &= \text{Concat}\left[C_{\text{i}} * \text{R}\left(\frac{k\pi}{2}\right)f\right]\\ &= \text{Concat}\left[\text{R}\left(\frac{k\pi}{2}\right)C_{\text{i}} * f\right], \quad k = 0, 1, 2, 3.
\end{aligned}
\end{equation}
This ensures alignment of ship center features while suppressing dispersed clutter.

To adjust the concatenated feature map $\widetilde{X} \in \mathbb{R}^{4C \times H \times W}$ to match the input dimensions, a $1 \times 1$ convolution is applied:
\begin{equation}
\text{Output} = \text{SiLU}\left(\text{BN}\left(\text{Conv}_{1 \times 1}(\widetilde{X})\right)\right),
\end{equation}
where $\text{BN}$ and $\text{SiLU}$ represent batch normalization and the sigmoid-weighted linear unit activation, respectively. 
By concatenating feature maps from diverse directions, the central regions of ships are reinforced, while clutter and speckle noise are suppressed. This design enhances targets' feature distinctiveness, improving localization in cluttered environments and ensuring robust target localization.

\begin{figure}[t]
\setlength{\belowcaptionskip}{-0.3cm}
  \begin{center}
  \includegraphics[width=0.9\linewidth]{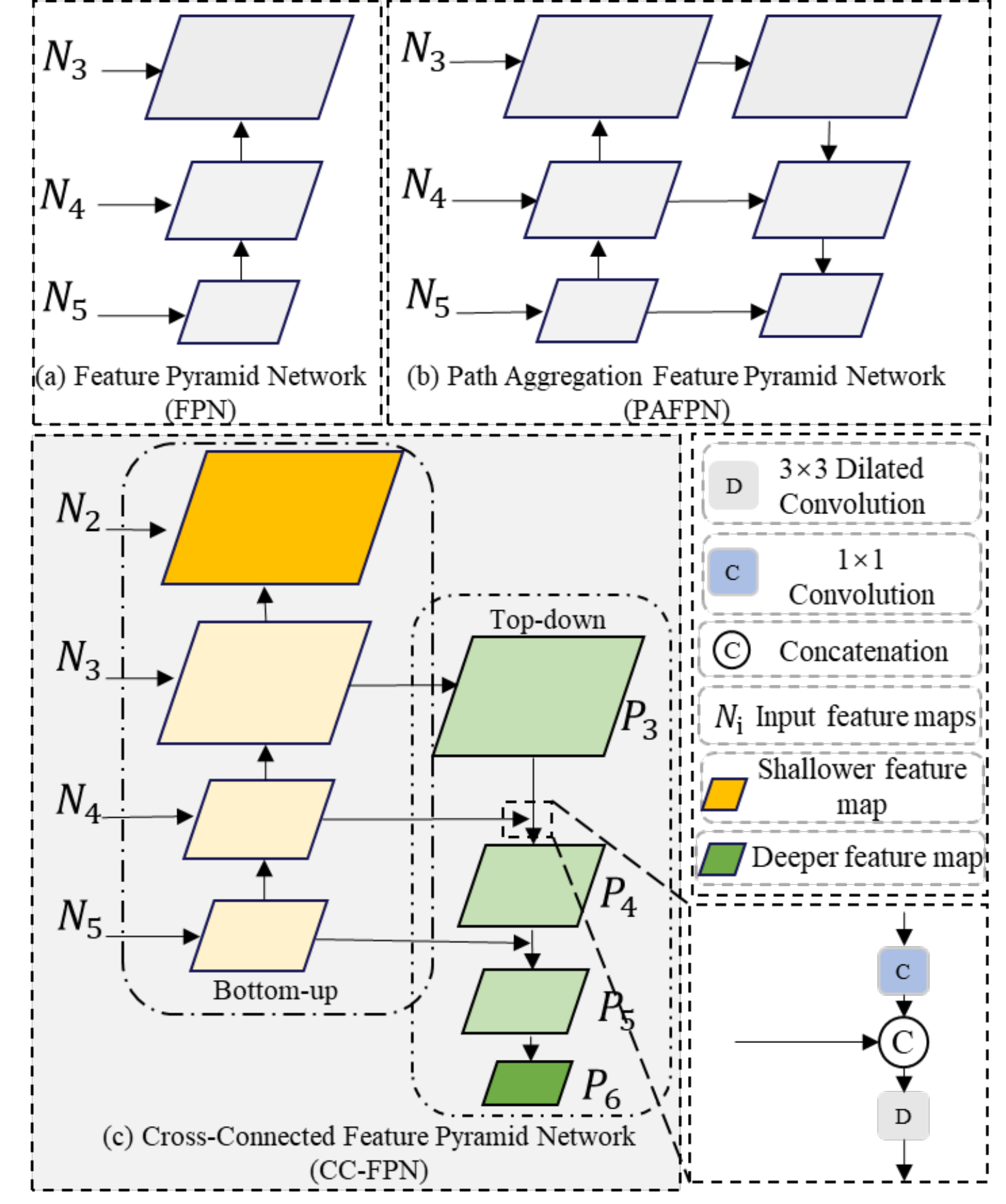}\\
  \captionsetup{font={footnotesize}}
  \caption{Structure of FPN, PAFPN, and CC-FPN (Ours). Compared with FPN and PAFPN, CC-FPN introduces extra shallow and deep feature maps with maintaining computation amount to enhance detail and semantic information, offering a richer portrayal of both small and large ships.}\label{CC-FPN}
  \end{center}
\end{figure}

\subsection{Neighbor Attention Module}
Inshore regions often exhibit densely packed ships, leading to overlapping or adjacent targets that are difficult to distinguish. While the CEM emphasizes ship centers, separating these closely located ships requires precise boundary refinement. 
The Neighbor Attention Module (NAM) achieves this by integrating different fine-grained global contextual information from adjacent levels. The specific structure of the NAM is shown in Fig.~\ref{sub} (b).

Given input feature maps $F_j \in \mathbb{R}^{C \times H \times W}$ and $F_{j+1} \in \mathbb{R}^{2C \times H/2 \times W/2}$ from adjacent layers, the NAM computes long-range dependencies using an attention mechanism. The query $Q$, key $K$, and value $V$ matrices are defined as:
\begin{equation}
\begin{aligned}
Q &= \text{Conv}_{1 \times 1}(F_j), \\
K &= \text{Conv}_{1 \times 1}(F_j), \\
V &= \text{Conv}_{1 \times 1}(\text{DeConv}_{3 \times 3}(F_{j+1})),
\end{aligned}
\end{equation}
where $\text{Conv}_{1 \times 1}$ and $\text{DeConv}_{3 \times 3}$ denote $1 \times 1$ convolution and $3 \times 3$ deconvolution, respectively.

The affinity matrix $A \in \mathbb{R}^{HW \times HW}$ is computed using the dot product between $Q$ and $K$:
\begin{equation}
A_{i,j} = \text{Softmax}(Q_i \cdot K_j^T),
\end{equation}
where $\text{Softmax}$ normalizes the affinities across spatial positions. The features are aggregated as:
\begin{equation}
H_i = \sum_j A_{i,j} \cdot V_j.
\end{equation}
The aggregated results are fused with the original inputs:
\begin{equation}
    F'_j = H + F_j, \quad F'_{j+1} = \text{Conv}_{3 \times 3}(H) + F_{j+1}.
\end{equation}
The outputs capture both local and global dependencies. 
Above processes for computing relationships between layers together form the neighbor cross attention (NCA). 
Single process produces the relationship between each pixel and the pixels in the row and column. Through iteratively refining these relationships, the NAM captures the global information of whole feature map. 
By combining different fine-grained global information from adjacent levels, NAM enhances the model's ability to separate overlapping targets and delineate precise boundaries, precisely positioning densely packed ships.

Meanwhile, an alternative configuration, In-NCA, reverses the resizing process to align shallow features with deeper ones, providing broader target-level correlations. Further experiments will evaluate the performance of NAM and In-NAM configurations.

\begin{figure*}[t]
\setlength{\belowcaptionskip}{-0.3cm}
 \begin{center}
  \includegraphics[width=0.95\linewidth]{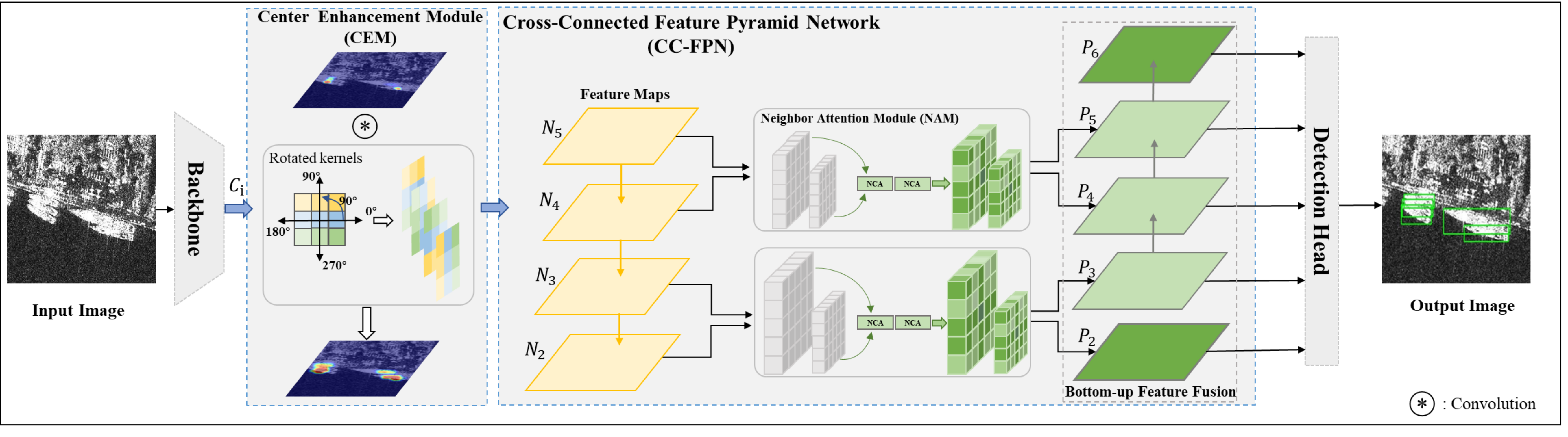}
\captionsetup{font={footnotesize}}
\caption{Overall architecture of CASS-Det. CEM and NAM represent center enhancement module and neighbor attention module, respectively. CASS-Det consists of the backbone network, four center enhancement modules, two neighbor attention modules, cross-connected feature pyramid network (CC-FPN), and detection head. Firstly, we adopt CEM to highlight the central regions of feature maps (\(C_i, (i=2,3,4,5)\)) based on the rotational convolution. Secondly, NAM is designed to combine various fine-grained global features by calculating the long-range dependencies based on the adjacent layers. Thirdly, CC-FPN provides richer semantic and contextual information through cross-connected feature fusion structure.  
}
\label{all_structure}
\end{center}
\end{figure*}

\subsection{Cross-Connected Feature Pyramid Network}

Ship sizes in SAR imagery vary significantly, ranging from small fishing boats to large cargo vessels. Traditional feature pyramid networks (FPN and PAFPN, as shown in Fig.~\ref{CC-FPN}(a) and (b)) often struggle to adapt to such extreme variations. The Cross-Connected Feature Pyramid Network (CC-FPN) addresses this limitation by integrating additional connections across feature levels for enhanced multi-scale feature fusion.

The CC-FPN fuses shallow detail features with deep semantic features. The detailed construction is shown in Fig. ~\ref{CC-FPN} (c). The shallowest feature map $N_2$ is directly connected to the detection head to preserve fine details, while deeper feature maps ($N_3$, $N_4$, and $N_5$) are refined using dilated convolutions:
\begin{equation}
P'_3 = \text{SiLU}\left(\text{BN}\left(\text{Conv}_{3 \times 3}(P_3)\right)\right),
\end{equation}
\begin{equation}
    P''_3=\text{Concat}(P'_3,N_4),
\end{equation}
\begin{equation}
P_4 = \text{SiLU}\left(\text{BN}\left(\text{DConv}_{1 \times 1}(P''_3)\right)\right),
\end{equation}
where $\text{DConv}_{3 \times 3}$ represents a $3 \times 3$ dilated convolution, which expands the receptive field to capture larger targets. The refined features are concatenated with higher-level feature maps for improved representation across scales. Moreover, a topmost feature map, $P_6$, is obtained through a $3\times 3$ convolution based on the $P_5$ to generate more global information. Compared with extending a level directly on the backbone, the feature extraction method reduces the fusion process and computation.
This cross-connected design enables the CC-FPN to enhance both shallow detail information and deep semantic information. By efficiently combining these features, CC-FPN handles both small and large targets effectively, achieving robust multi-scale detection while maintaining computation amount.

\subsection{Integration into the Existing Framework}
The proposed modules are integrated into a unified framework built upon CSPDarknet as the backbone and YOLOX as the detection head. The backbone generates multi-level feature maps, which are processed by the CEM to enhance ship features and suppress background noise. The CC-FPN then fuses these features across scales, while the NAM refines boundaries of densely packed ships.
Finally, the YOLOX detection head synthesizes the processed features to generate bounding boxes and classifications. The network employs IoU loss as the bounding box regression loss, and calculates the classification loss and confidence loss through Binary Cross-Entropy Loss. This modular design improves detection accuracy for multi-scale and densely packed ships, making the proposed framework both effective and efficient for SAR-based maritime applications.

\begin{table*}
    \renewcommand{\arraystretch}{1.2}
    \centering
    \caption{\normalsize{C}\footnotesize{OMPARISON} \normalsize{O}\footnotesize{F} \normalsize{D}\footnotesize{IFFERENT} \normalsize{M}\footnotesize{ETHODS} \normalsize{O}\footnotesize{N} \normalsize{S}\footnotesize{SDD.}}
    \begin{tabular}{c|c|c|c|ccccccc}
    \Xhline{1.2pt}
        \multicolumn{2}{c|}{Method}& Backbone& Neck& mAP& F1& Recall& Precision& \(\text{AP}_\text{s}\)&  \(\text{AP}_\text{m}\)&  \(\text{AP}_\text{l}\)\\ 
    \hline
        \multirow{4}*{Two-stage}& SER Faster R-CNN \cite{SER-Faster}& VGG & -- &0.915 & 0.891 &0.923 & 0.861 &  -- & -- & --\\    
        ~&Dynamic R-CNN\cite{Dynamic} & ResNet-50 & FPN & 0.922& 0.908& 0.903& 0.913 & 0.544&  0.627& 0.541\\
        ~&Double-head R-CNN\cite{Double-head} & ResNet-50 & FPN &  0.925& 0.915& 0.899& 0.930 & 0.529&  0.599& 0.520\\
        ~&Cascade R-CNN\cite{22} & ResNet-101 & FPN &0.941& 0.936& 0.930& 0.941 & 0.547&  0.642& 0.661\\
    \hline
        \multirow{10}*{One-stage} &RetinaNet\cite{29} & ResNet-101 & FPN & 0.880& 0.862& 0.824& 0.903 &   0.500&  0.638& 0.702\\
        ~&CR2A-Net\cite{CR2A-Net} & ResNet-101 & FPN & 0.898 &0.907 & 0.878 &0.940 &  --& --&--\\
        ~&FCOS\cite{FCOS} & ResNet-101 & FPN & 0.937& 0.925& 0.901& 0.950 &   0.547&  0.688& 0.714\\
        ~&Free-anchor\cite{Free-anchor} & ResNet-101 & FPN & 0.939& 0.924&  0.930& 0.917 &  0.545&  0.644& 0.680\\
        ~&YOLO-FA\cite{YOLO-FA} & CSPDarknet & PAFPN & 0.968 & 0.950 & 0.950 &0.952 & --& --&--\\
        ~&YOLO v8\cite{Yolov8} & CSPDarknet & PAFPN & 0.970& 0.948& 0.923& 0.967 & 0.489&  0.684& 0.665\\
        ~&YOLOX\cite{Yolox}& CSPDarknet & PAFPN & 0.977& 0.946& 0.928& 0.965&   0.543&  0.663& 0.634\\
        ~&BiFF-FESA\cite{BiFF-FESA} & CSPDarknet & PAFPN & 0.978& 0.950 &  0.940 & 0.961 &--& --&--\\    
        ~&YOLO v7\cite{Yolov7} & CSPDarknet & PAFPN & 0.980& 0.943& 0.919& 0.967 &    0.502&  0.680& 0.711\\
    \hline
        ~& Ours& CSPDarknet & CC-FPN & \textbf{0.992}& \textbf{0.974} & \textbf{0.978}& \textbf{0.969} & \textbf{0.568}& \textbf{0.706} & \textbf{0.737}\\
    \Xhline{1.2pt}
    \end{tabular}
    \label{tab:SSDD-con}
\end{table*}

\begin{table*}
    \renewcommand{\arraystretch}{1.2}
    \centering
    \caption{\normalsize{C}\footnotesize{OMPARISON} \normalsize{O}\footnotesize{F} \normalsize{D}\footnotesize{IFFERENT} \normalsize{M}\footnotesize{ETHODS} \normalsize{O}\footnotesize{N} \normalsize{H}\footnotesize{RSID.}}
    \begin{tabular}{c|c|c|c|ccccccc}
    \Xhline{1.2pt}
         \multicolumn{2}{c|}{Method}& Backbone& Neck& mAP& F1& Recall& Precision&  \(\text{AP}_\text{s}\)&  \(\text{AP}_\text{m}\)&  \(\text{AP}_\text{l}\)\\ 
    \hline
        \multirow{5}*{Two-stage}&Dynamic R-CNN\cite{Dynamic} & ResNet-50 & FPN  & 0.830& 0.851& 0.836& 0.866 &   0.454&  0.735& 0.350\\      
        ~&Cascade R-CNN\cite{22} & ResNet-101 & FPN & 0.848& 0.862& 0.855& 0.870 & 0.458&  0.731& 0.409\\ 
        ~&Double-head R-CNN\cite{Double-head} & ResNet-50 & FPN & 0.852&  0.854&  0.868& 0.840 &  0.465&  0.733& 0.359\\
        ~&ATSD\cite{ATSD} & DAL-34 & WCFF-FPN & 0.881 & 0.883 & 0.865 & 0.902 &  -- &--&--\\
    \hline
        \multirow{9}*{One-stage}&RetinaNet\cite{29} & ResNet-101 & FPN & 0.798&   0.777& 0.757& 0.797 &  0.343&  0.728& 0.442\\
        ~&Free-anchor\cite{Free-anchor} & ResNet-101 & FPN & 0.835& 0.837& 0.828&  0.846 & 0.413&  0.728& 0.463\\
        ~&FCOS\cite{FCOS} & ResNet-101 & FPN &  0.867& 0.852&  0.821 & 0.887 &  0.424&  0.730& 0.449\\
        ~&DWB-YOLO\cite{DWB-YOLO} & -- & -- & 0.888 &0.850 & 0.805 &  0.900 &-- & --& --\\
        ~&YOLO v7\cite{Yolov7} & CSPDarknet & PAFPN & 0.911& 0.877&  0.863&  0.891 &  0.384&  0.718& 0.422\\
        ~&YOLO v8\cite{Yolov8} & CSPDarknet & PAFPN & 0.913& 0.872&  0.845&  0.901 &  0.460&  0.726 & 0.407\\
    \hline
        ~&Ours & CSPDarknet & CC-FPN & \textbf{0.931} & \textbf{0.901} &\textbf{0.881}& \textbf{0.922} &  \textbf{0.476} &  \textbf{0.736}& \textbf{0.468}\\
    \Xhline{1.2pt}
    \end{tabular}
    \label{tab:HRSID-con}
\end{table*}

\begin{figure*}[t]
\setlength{\belowcaptionskip}{-0.3cm}
  \centering
  \includegraphics[width=0.95\linewidth]{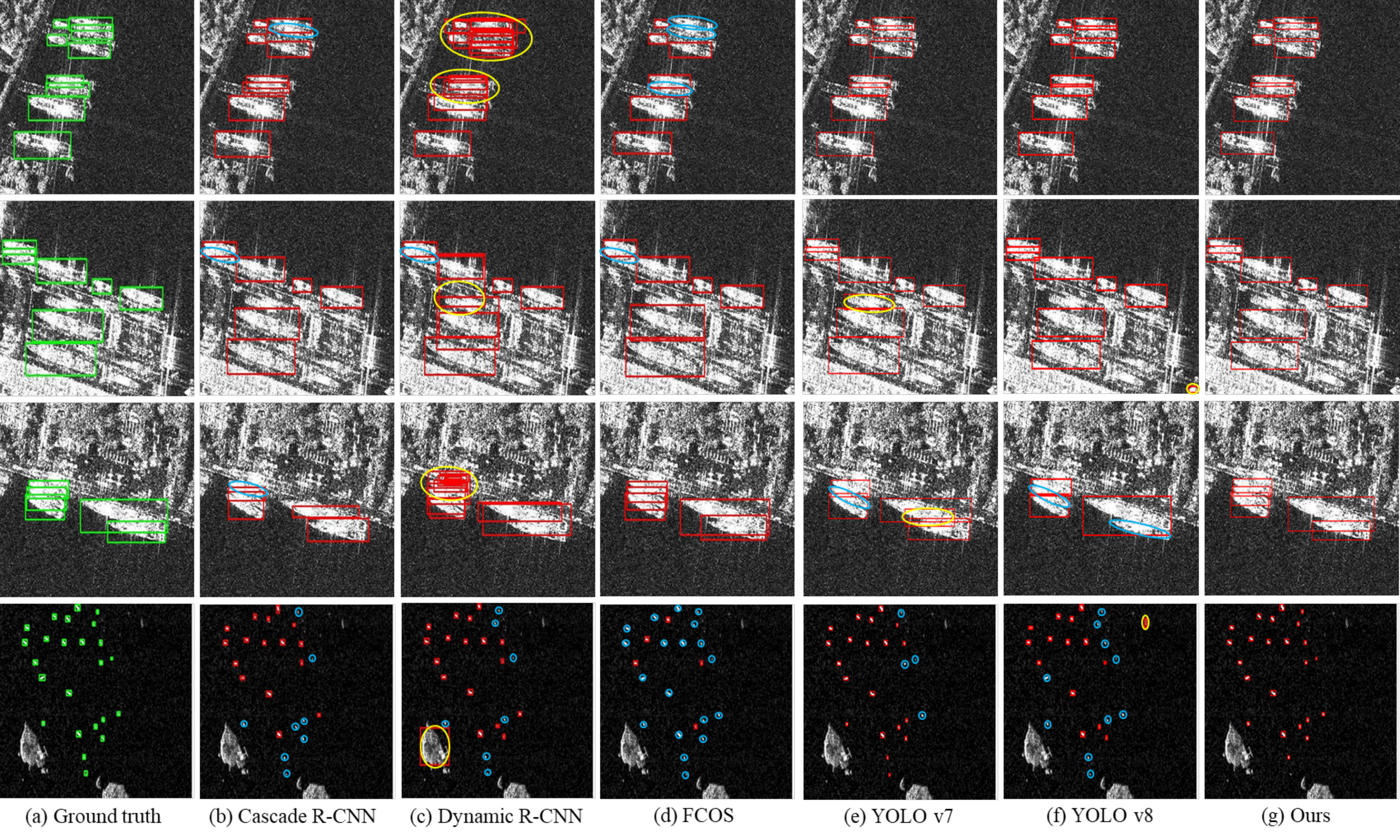}
  \captionsetup{font={footnotesize}}
\caption{Detection Results of Different Methods on SSDD. The green box is the ground truth. Red box is the detection result. The yellow circle indicates false detection, and the blue circle indicates missed detection.}
\label{SSDD}
\end{figure*}

\begin{figure*}[t]
  \centering
  \includegraphics[width=0.95\linewidth]{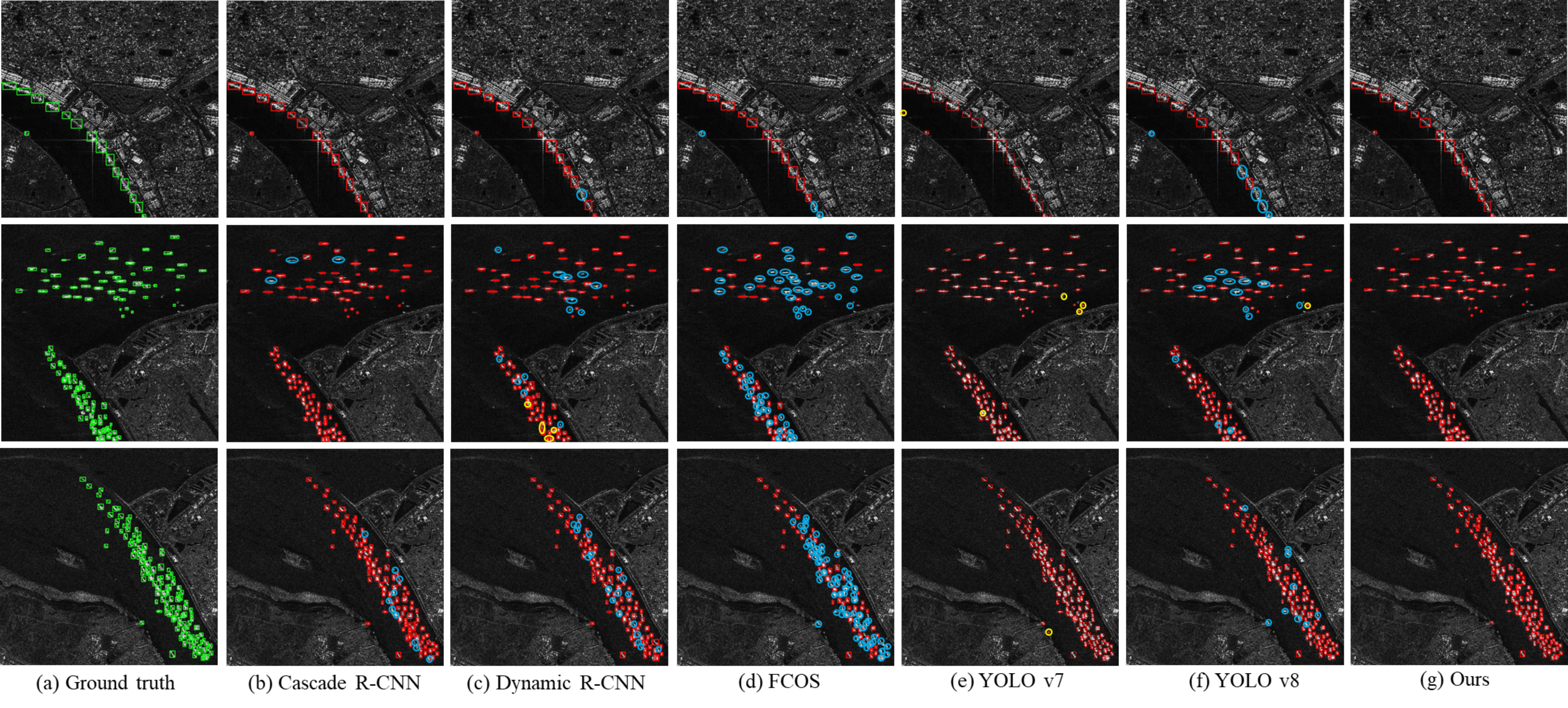}
  \captionsetup{font={footnotesize}}
\caption{Detection Results of Different Methods on HRSID. The green box is the ground truth. Red box is the detection result. The yellow circle indicates false detection, and the blue circle indicates missed detection.}
\label{HRSID}
\end{figure*}

\section{Experiments}

In this section, experimental details and results will be introduced. First, datasets and parameter settings will be presented in detail. Then, we will conduct ablation studies to validate the effectiveness of each component of the proposed method. Finally, the comparison with other methods will be shown in Section III-C.

\subsection{Implementation Details}
\subsubsection{Experimental datasets}
To fully verify the validity and universality of the model, we use three multi-resolution, multi-scene, multi-polarization data sets: SSDD, HRSID, and LS-SSDD for experiments.

SSDD is mainly from RadarSat-2, TerraSAR-X, and Sentinel-1 detectors, with resolutions ranging from 1m to 15m and polarization modes of HH, HV, VH. It contains 1160 SAR images and 2456 multi-scale ships, the average of which is 2.12 ships per image. It contains different scenes, including inshore, offshore, and multi-obstacle. Therefore, it can be taken to verify the stability and universality of the methods. The sizes of images are about 500×500. We unify the sizes as 512×512 in the experiments. According to the official recommendations, we take images’ suffixes 1 and 9 as the test set, 8 as the verification set, and the rest as the training set.

Compared with SSDD, HRSID has larger scenes and more small ships, bringing more difficulty to accurate detection. The images are from Sentinel-1B, TerraSAR-X, and TanDEMX sensors. It includes 5604 images with 16951 ships with 800×800 pixel resolutions. The polarization modes contain HH, HV, and VV. The resolution of images ranges from 0.5 to 3m. We divide the training and test sets as 0.65:0.35 according to the given contributions.

LS-SSDD utilizes Sentinel-1 satellite data and includes a total of 30 large-scale SAR images whose sizes are 24000×16000. The polarization modes including VV and VH, and the imaging mode is IW. The dataset is characterized by large-scale maritime observation, small-scale ship detection, a variety of pure backgrounds, a fully automated detection process, and multiple standardized benchmarks. In the experiments, we split each of the images into 800×800 sub-images for training and testing convenience. We take 6000 sub-images for training and 600 for testing. 

\subsubsection{Settings}

Experiments in Section III are all implemented by Pytorch and operated on Nvidia GeForce RTX 3090 GPUs. We set 500 epochs for training based on transfer learning. Furthermore, Adam is set as the optimizer and the batchsize is 8.

\subsection{Evaluation Criteria}

To evaluate the performance of method, we employ mAP, precision, recall, F1 as evaluate criteria. To comprehensively evaluate the detection performance of the proposed method on ships of different scales, we adopt the evaluation method in the COCO dataset to obtain \(\text{AP}_\text{s}\), \(\text{AP}_\text{m}\), and \(\text{AP}_\text{l}\), which represent the detection accuracy of small(\(\text{area} \textless 32^2 \text{pixels}\)), medium(\(32^2 \textless \text{area} \textless 96^2 \text{pixels}\)), and large ships(\(\text{area} \textgreater 96^2\)), respectively.

\subsection{Comparison Experiment}

To demonstrate the superior properties of our method, we compare CASS-Det with the existing methods on the SSDD, HRSID and LS-SSDD. 
 The comparison methods include one-stage methods (FCOS, RetinaNet, Free-anchor, YOLO v7, YOLO v8, CR2A-Net\cite{CR2A-Net}, YOLO-FA\cite{YOLO-FA}, and BiFF-FESA\cite{BiFF-FESA}) and two-stage methods (Cascade R-CNN, Double-head R-CNN, Dynamic R-CNN, ATSD\cite{ATSD}, and SER Faster R-CNN\cite{SER-Faster}). The results are shown in Table \ref{tab:SSDD-con} and Table \ref{tab:HRSID-con}. 

\noindent\textbf{1) SSDD: }The detection performance on SSDD is presented in Table \ref{tab:SSDD-con}. Compared with two-stage methods, one-stage ones have better performance. CASS-Det achieves 0.992 mAP, 1.2$\%$-11.2$\%$ higher than other methods. It has the highest recall, 0.978, signifying it has the fewest missed detections. YOLO v7 is ranked as the second-best performing detector among comparison methods. CASS-Det is 1.2\% higher than that of YOLO v7. 
Although the precision of CASS-Det is only 0.2\% higher than that of YOLO v8 and YOLO v7, it performs well in pinpointing each target, providing 5.5\% and 5.9\% advantages in recall. This is because CASS-Det emphasizes the center of targets and refines ship boundaries based on the CEM and NAM. Furthermore, CASS-Det has the most significant advantage in detecting all-scale ships, with 1.8$\%$-2.3$\%$ promotions compared with the second-best methods. 

To obtain the visual effect of detection, we selected images from SSDD to test, and the results are shown in Fig.~\ref{SSDD}. The green rectangles in (a) are the ground truth, and red rectangles in (b)-(g) are the detection results of different methods. The yellow and blue circles indicate false and missed detection, respectively. 
Compared with the other methods, CASS-Det generates less false and missed detections in various scenes, including densely arranged ships (first row), inshore interferences (second and third rows), and small ships (fourth row). YOLO v7 and YOLO v8 perform well in detecting densely arranged ships, but have some missed detections in inshore and small-ship scenes. 
FCOS can detect ships in the inshore scenes well, but have missed detections in densely arranged and small-ship scenes. Benifiting from the CEM and NAM, the CASS-Det has a better sense of the ships, locating each ship precisely. It proves that CASS-Det is adaptable to complex environments. In summary, CASS-Det has achieved state-of-the-art performance on the SSDD.

\noindent\textbf{2) HRSID: }Table \ref{tab:HRSID-con} shows the results of comparative experiments on the HRSID. Compared to the dataset mentioned above, the HRSID contains more intricate scenes and a greater number of small ships. This is intended to better reflect the effectiveness and stability of the methods utilized. 
CASS-Det achieves 0.931 mAP on the HRSID benefiting from CEM, NAM, and CC-FPN. CEM leads attention to the center of the ship. NAM fuses long-range dependencies of different grains to enhance global features and accurately pinpoints the boundaries of ships. CC-FPN brings additional shallow and deep feature maps, enhancing the network's perception of detailed and semantic features. 
Moreover, recall, precision, and F1 of the method achieve perfect, verifying the balance of the method. YOLO v8 is ranked as the second-best detector among comparison experiments. CASS-Det achieves 1.8\%, 2.1\%, and 3.6\% higher than YOLO v8 on the mAP, precision, and recall, respectively. The precision and recall increase uniformly, improving F1 and mAP. 
Moreover, the \(\text{AP}_\text{l}\), \(\text{AP}_\text{m}\), and \(\text{AP}_\text{s}\) are optimal with 0.476, 0.736, and 0.468. Although small ships account for the largest proportion in the HRSID, CASS-Det has the best performance. 

In the Fig.~\ref{HRSID}, land occupies most space of the image, making the interferences complex. As the large-scene and high-resolution imaging, the size of each ship is small, increasing the detection difficulty. YOLO v7 produces few false alarms in the inshore and small-ship scenes as the complex land interferences producing false alarms. Other contrast methods, including Cascade R-CNN, Dynamic R-CNN, FCOS, and YOLO v8 have many missed detections and a few false alarms. Meanwhile, CASS-Det outperforms other methods in both scenes. With the CASS-Det's effect of concentrating on the central parts of ships, CASS-Det can pinpoint the location of each ship, leading to the accurate detection effect. Both quantitative metrics and visual outcomes demonstrate the superior performance of CASS-Det.

\noindent\textbf{3) LS-SSDD: }To assess the generalization capabilities of our method, we employ large-scale scene images for validation. Compared to SSDD and HRSID, the LS-SSDD dataset presents larger images and smaller ships, posing a more significant detection challenge. 
The evaluation metrics are summarized in Table.\ref{tab:LS-SSDD}. Our proposed method attains a precision of 0.911 and an mAP of 0.821, outperforming other comparison methods by up to 32.7\% in precision and 11.2\% in mAP. 
Compared with YOLOX, the metrics achieve better performance with the benefit of three modules. mAP and precision are 1.8\% and 1.1\% higher than YOLOX, verifying the effective of modules.

\begin{table}[t]
    \renewcommand{\arraystretch}{1.2}
    \begin{center}
    \caption{\centering{\normalsize{C}\footnotesize{OMPARISON} \normalsize{O}\footnotesize{F} \normalsize{D}\footnotesize{IFFERENT} \normalsize{M}\footnotesize{ETHODS} \normalsize{O}\footnotesize{N} \normalsize{L}\footnotesize{S-SSDD.}}}
    \begin{tabular}{c|c|cc}
    \Xhline{1.2pt}
         \multicolumn{2}{c|}{Method}&mAP& Precision\\
    \hline
         \multirow{5}*{Two-stage} &Cascade R-CNN\cite{22}&0.709& 0.841\\
         ~&Libra R-CNN\cite{45}&0.737&  0.736\\
         ~&DCN\cite{DCN}&0.738& 0.741\\
         ~&Faster R-CNN\cite{20}&0.748 &0.737\\
         ~&MSIF\cite{MSIF}&0.756& 0.866\\
    \hline
         \multirow{5}*{One-stage}&L-YOLO\cite{L-YOLO}&0.730&0.848\\
         ~&SII-Net\cite{SII-Net} & 0.761&0.682\\
         ~&SAR-Net\cite{SAR-Net}&0.762& 0.584\\
         ~&YOLOX\cite{Yolox}&0.803& 0.900\\
    \hline
         ~&Ours&\textbf{0.821}& \textbf{0.911}\\
    \Xhline{1.2pt}
    \end{tabular}
    \label{tab:LS-SSDD}
    \end{center}
\end{table}

\begin{figure*}[t]
\setlength{\belowcaptionskip}{-0.01cm}
  \centering
  \includegraphics[width=0.95\linewidth]{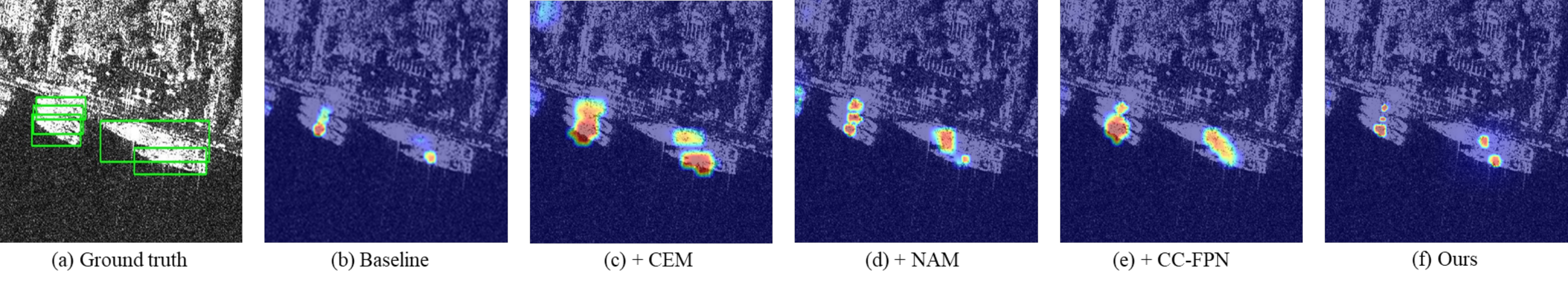}
  \captionsetup{font={footnotesize}}
\caption{Heatmaps of networks with different modules. Green boxes are ground truth.}
\label{hm}
\end{figure*}

\begin{figure*}[t]
\setlength{\belowcaptionskip}{-0.3cm}
  \centering
  \includegraphics[width=0.95\linewidth]{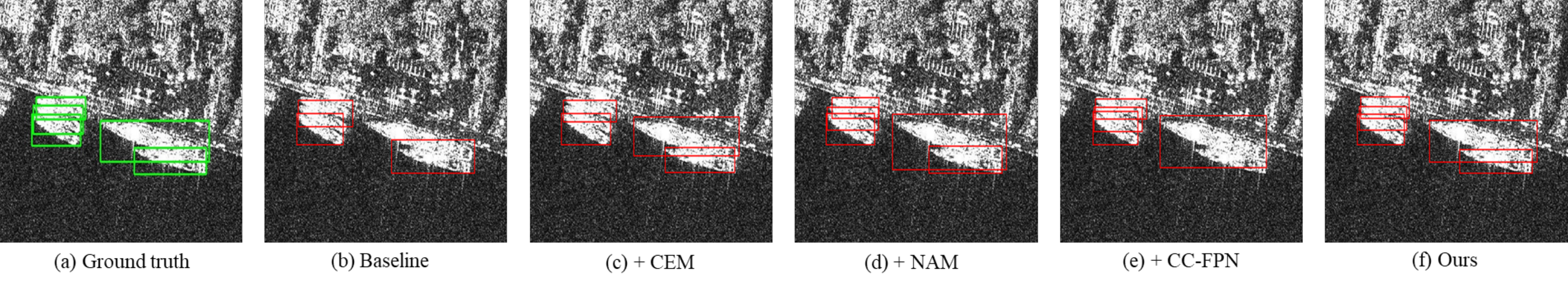}
  \captionsetup{font={footnotesize}}
\caption{Detection results of different modules. Green boxes are ground truth. Red boxes are the detection results.}
\label{res8}
\end{figure*}

\begin{table*}[t]
    \renewcommand{\arraystretch}{1.5}
    \centering
    \caption{\normalsize{A}\footnotesize{BLATION} \normalsize{S}\footnotesize{TUDIES} \normalsize{O}\footnotesize{F} \normalsize{C}\footnotesize{ASS-DET.} }
    \begin{tabular}{cccc|ccccccc}
    \Xhline{1.5pt}
         Baseline& CEM& NAM &CC-FPN & mAP& F1 & Recall & Precision & \(\text{AP}_\text{s}\)&   \(\text{AP}_\text{m}\)&  \(\text{AP}_\text{l}\)\\
    \hline
        \checkmark & \(\times\) & \(\times\) &\(\times\) &  0.977 & 0.946 & 0.928 & 0.965&  0.543&  0.663& 0.634\\
        \checkmark & \checkmark & \(\times\) & \(\times\)  & 0.982 & 0.954 & 0.941 &0.966 &  0.550&  0.696& 0.670\\
        \checkmark & \(\times\) & \checkmark & \(\times\)  & 0.988 & 0.964 &0.969 &0.960&  0.566&  0.699& 0.723\\
        \checkmark & \(\times\) & \(\times\) & \checkmark & 0.980 & 0.954 & 0.940 &0.968& 0.544&  0.670& 0.692\\
        \checkmark & \checkmark & \(\times\) &\checkmark & 0.985 & 0.956 & 0.945 & 0.968 &  0.556&  0.695& 0.721\\
        \checkmark & \checkmark &\checkmark & \checkmark &  \textbf{0.992}&\textbf{0.974} & \textbf{0.978}& \textbf{0.969} &  \textbf{0.568}& \textbf{0.706} & \textbf{0.737}\\
    \Xhline{1.2pt}
    \end{tabular}
    \label{tab:abl}
\end{table*}

\begin{table*}[t]
    \renewcommand{\arraystretch}{1.5}
    \centering
    \caption{\normalsize{E}\footnotesize{XPERIMENTAL} \normalsize{R}\footnotesize{ESULTS} \normalsize{O}\footnotesize{F} \normalsize{F}\footnotesize{PN,} \normalsize{P}\footnotesize{AFPN,} \normalsize{A}\footnotesize{ND} \normalsize{C}\footnotesize{C-FPN.}}
    \begin{tabular}{c|ccc|ccccccc}
    \Xhline{1.5pt}
          Backbone& FPN & PAFPN& CC-FPN&mAP&  F1&  Recall& Precision&\(\text{AP}_\text{s}\)&  \(\text{AP}_\text{m}\)&  \(\text{AP}_\text{l}\)\\ 
    \hline
        \multirow{3}*{CSPDarknet}& \checkmark & \(\times\) & \(\times\) &0.970 & 0.925 &0.947 &0.904 & 0.570&  0.690& 0.657\\
        ~& \(\times\)  & \checkmark & \(\times\) & 0.977 & 0.946 & 0.928 & 0.965& 0.543&  0.663& 0.634\\
        ~& \(\times\) &\(\times\) & \checkmark & 0.980 & 0.954 & 0.940 & \textbf{0.968}&  0.544&  0.670& 0.692\\
    \hline
        \multirow{2}*{CSPDarknet+CEM}&\(\times\) & \checkmark &\(\times\) &  0.982&  0.954&  0.941& 0.966& 0.550&  \textbf{0.696}& 0.670\\
        ~ & \(\times\) & \(\times\) &\checkmark & \textbf{0.985} &  \textbf{0.956}& \textbf{0.945} & \textbf{0.968}&  \textbf{0.556}&  0.695& \textbf{0.721}\\
    \Xhline{1.2pt}
    \end{tabular}
    \label{tab:CC-FPN}
\end{table*}

\subsection{Ablation Studies and Analysis}
We verify the effect of different modules (CEM, NAM, and CC-Net) in the CASS-Det through ablation studies. The results are summarized in Table ~\ref{tab:abl}, Fig.~\ref{hm}, and Fig.~\ref{res8}. We take the combination of CSPDarknet and FPN as the baseline. 

\noindent\textbf{(1) Effect of CEM: }Compared with the baseline, the CEM promotes mAP by 0.5\%. CEM leads the model's concentration to the central regions of targets through the intersection of features in different directions. The highlight of central area identifies the location of each target, elevating recall from 92.8\% to 94.1\%, with a promotion of 1.3\%. Meanwhile, the precision also increases 0.1\%, making recall and precision more balance. 
For the detection results of each scale targets, CEM significantly enhances the detection accuracy of models for medium and large-sized targets with 3.3\% and 3.6\% promotion, respectively. This is due to the fact that the central areas of medium to large-sized targets possess larger intersecting zones, thereby making them more prone to capturing the model's attention.
With the CC-FPN added, recall achieves an improvement by 0.5\%. The accuracy of small, medium, and large targets increase by 1.2\%, 2.5\%, and 1.9\%. The improvements further verify the effect of CEM.

\noindent\textbf{(2) Effect of NAM: }Through the analysis of NAM and In-NAM, we adopt the NAM as the long-range dependencies calculation method. From Table ~\ref{tab:abl}, although NAM descends 0.5\%, recall, F1, and mAP increase by 4.1\%, 1.8\%, and 1.1\%. 
With the better feature perception effect of all granularities from NAM, all scales of targets achieve more accurate detection effect. \(\text{AP}_\text{s}\), \(\text{AP}_\text{m}\), and \(\text{AP}_\text{l}\) increase by 2.3\%, 3.2\%, and 8.9\%. This proves the effectiveness of the combination of long-range denpendencies from different layers.
With CEM and CC-FPN added, NAM also has contributions to all metrics, with the improvement of precision, recall, F1, and mAP 0.1\%, 3.3\%, 1.8\%, and 0.7\%, respectively. In addition, the accuracy of all scales ships. It indicates the necessary to introduce NAM into the network.

\noindent\textbf{(3) Influence of CC-FPN: }From the results in Table ~\ref{tab:abl}, CC-FPN results in better precision and recall with promotion of 0.3\% and 1.2\% compared with the baseline. With these increase, mAP and F1 raise from 0.977 and 0.946 to 0.980 and 0.954. It shows that all indicates are significantly raised with CC-FPN.
CC-FPN adds deep and shallow feature extraction layers in the module, bringing better detailed and semantic information to the network. Correspondingly, \(\text{AP}_\text{s}\), \(\text{AP}_\text{m}\), and \(\text{AP}_\text{l}\) increase by 0.1\%, 0.7\%, and 5.8\%, indicating the forward gain of CC-FPN.
With the CEM added, precision, recall, F1, and mAP raise by 0.2\%, 0.4\%, 0.2\%, and 0.3\%, respctively. Although \(\text{AP}_\text{m}\) has a decline of 0.1\%, \(\text{AP}_\text{s}\) and \(\text{AP}_\text{l}\) increase by 0.6\% and 5.1\%, verifying the effectiveness of CC-FPN.

\noindent\textbf{(4) Visual Comparisons: }Fig.~\ref{hm} and Fig.~\ref{res8} show the heatmaps and detection results of network with different modules. As shown in Fig.~\ref{hm}(b), the baseline network can only perceive the central regions of small ships, missing the large ship. The centers of two adjacent small ships merged, resulting in the oversight of one of the small ships. 
CEM focus on the central parts of targets through the intersection of rotational convolution. Correspondingly, CEM extracts more central areas compared with the baseline. The large ship is detected accurately. However, the three adjacent small ships are mixed up, resulting in a missed detection of the small ship. 
NAM introduces long-range dependencies of different granularities into the network, distinguishing the adjacent ships from each other. From Fig.~\ref{hm}(d), it can be seen that the highlighted areas of different ships are separate, but the central part of the large ship is spread. As a result, the detection bounding box of the large ship is excessively large. 
CC-FPN improves the feature extraction ability of the network for large and small targets, but it cannot separate the densely arranged targets, resulting in the missed detection of two ships as one ship in the detection result. 
As shown in Fig.~\ref{hm}(f) and Fig.~\ref{res8}(f), by integrating the above modules, the network centers its attention on the core area of each ship, ensuring that this central region does not surpass the width of the ship. This allows the algorithm to precisely pinpoint the location and delineate the boundary of each target.

\noindent\textbf{(5) Analysis of NAM and In-NAM: }The combination of long-range features with different granularities helps the network to distinguish the densely-arranged inshore ships from each other and define the ship's boundaries. To verify the rationality of the model, we compare NCA and In-NCA in this section. The difference between NAM and In-NAM is the long-range features’ different fine-grained. Compared to the In-NAM, NAM operates on a shallower feature map, so the extracted features have finer granularities. We conducted the experiments on the SSDD based on the CSPDarknet with CEM and PBC-FPN. The results are presented in Fig.~\ref{NAM_res}. Although the \(\text{AP}_\text{m}\) of NAM is 0.6$\%$ lower than that of In-NAM, the  \(\text{AP}_\text{s}\) and  \(\text{AP}_\text{l}\) of NAM are 1.1$\%$ and 3.7$\%$ higher than those of In-NAM. NAM is superior to In-NAM in detecting small and large ships as NAM has finer granularities, bringing a better perception of targets. Therefore, NAM is employed in CASS-Det as an enhanced module.

\begin{figure}[t]
\setlength{\belowcaptionskip}{-0.1cm}
  \centering
  \includegraphics[width=0.95\linewidth]{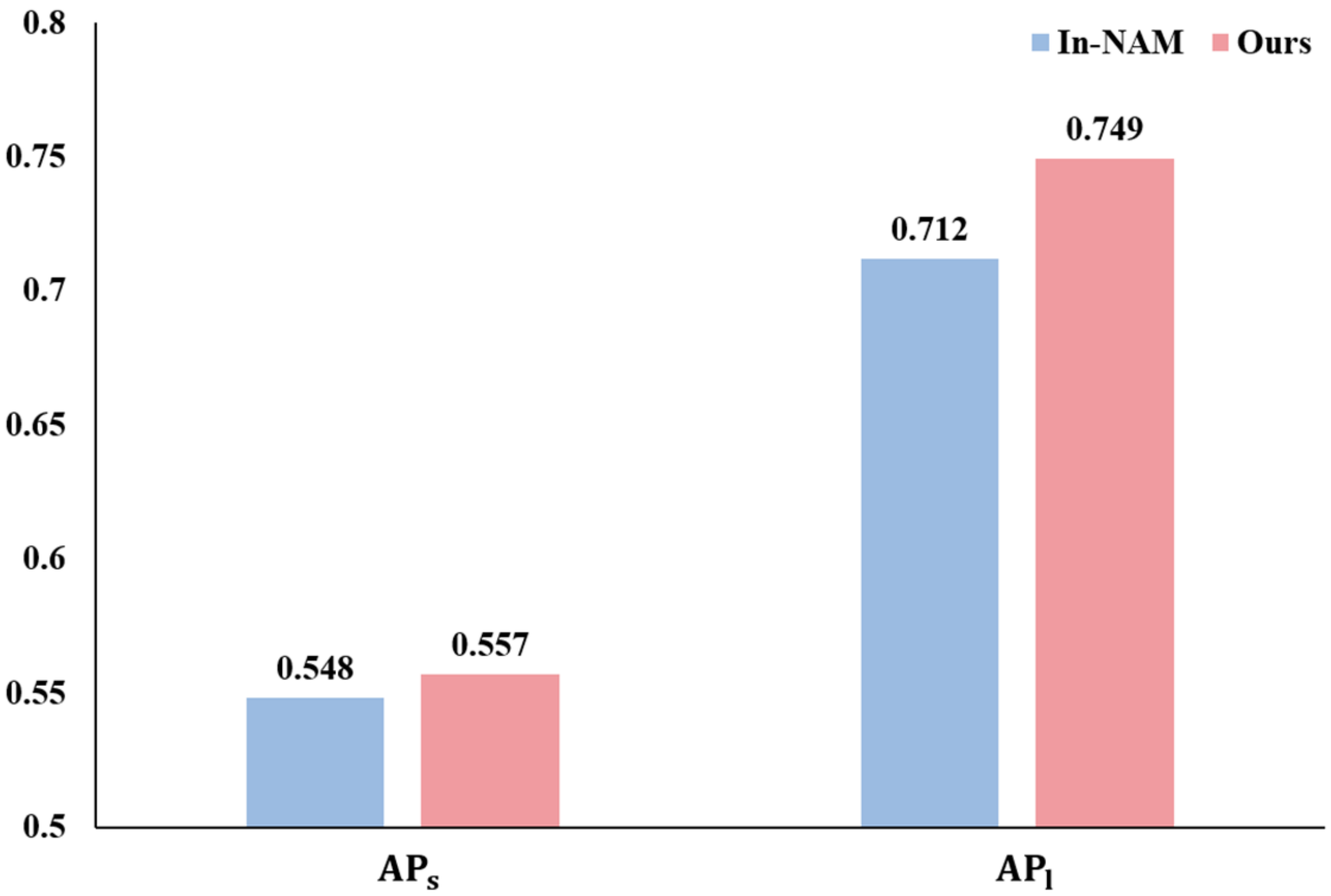}
  \captionsetup{font={footnotesize}}
\caption{Experimental results of NAM and In-NAM.}
\label{NAM_res}
\end{figure}

\noindent\textbf{(6) Analysis of FPN, PAFPN, and CC-FPN: }To assess the performance of CC-FPN, we conducted experiments on the SSDD. The detection results are presented in Table.\ref{tab:CC-FPN}. It can be observed from Table.\ref{tab:CC-FPN} that, in comparison with the baseline and PAFPN, CC-FPN demonstrates enhancement in detection performance across all cases. Comparing with the baseline, CC-FPN improves the precision and mAP with 6.4\% and 1.0\%, respectively. Comparing with the PAFPN, CC-FPN increases the recall by 0.3$\%$, with \(\text{AP}_\text{s}\), \(\text{AP}_\text{m}\) and \(\text{AP}_\text{l}\) 0.1$\%$, 0.7$\%$, and 5.8$\%$, respectively. With the CEM added, CC-FPN promotes all indexes, with mAP,  \(\text{AP}_\text{s}\) and  \(\text{AP}_\text{l}\) 0.3$\%$, 0.6$\%$, and 5.1$\%$, respectively. It shows that CC-FPN can reduce false alarms better than PAFPN, improving the detection effect on ships of different sizes, especially large ones. This is because the CC-FPN increases the number of feature levels, expands the receptive field, and enhances the detection ability of the network for multi-scale ships.

\noindent\textbf{(7) Combination with Other Framework: }To verify the universality of the modules, we verify the CEM, NAM, and CC-FPN under the frameworks of YOLO v7 and YOLO v8 on the SSDD. The recalls of original methods and those combined with three models are shown in Fig.~\ref{recall}. With three modules added, recalls of three methods elevate from 4.3\% to 5\%. Benefitted from the center enhancement effect of CEM and NAM, targets are more distinguishable and accurately determined. CC-FPN promotes the  model's ability to perceive large and small targets. The combination of three modules enhances the detection effect.

\begin{figure}[t]
\setlength{\belowcaptionskip}{-0.1cm}
  \centering
  \includegraphics[width=0.95\linewidth]{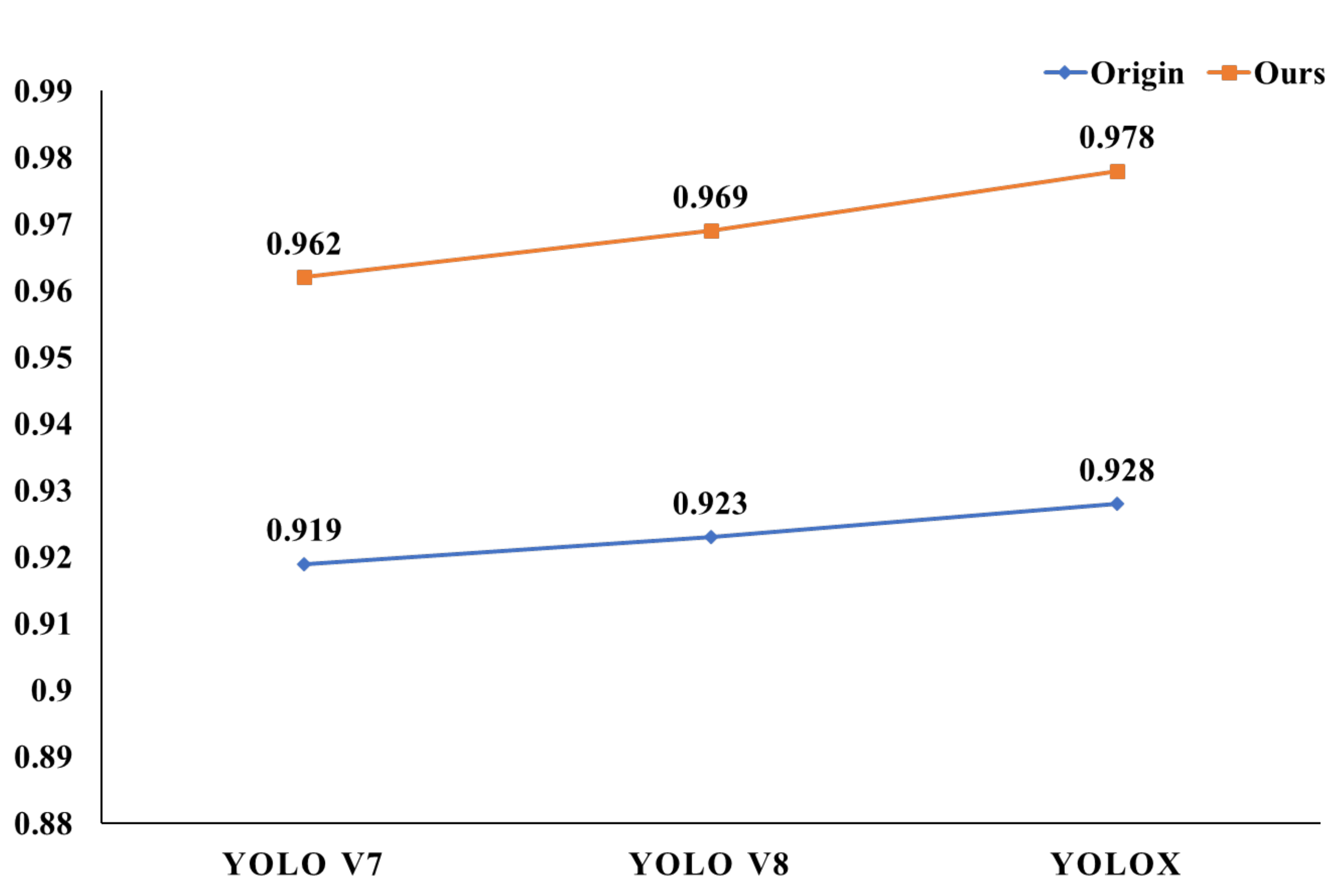}
  \captionsetup{font={footnotesize}}
\caption{Recalls of original YOLOs and those combined with our medules.}
\label{recall}
\end{figure}

\section{Conclusion}

This paper proposes a novel, robust SAR ship detection method, CASS-Det. 
To address the intricate backgrounds prevalent in SAR images, the CEM is devised. It can extract multi-direction ship features through rotational convolution. With the overlaping of feature maps, the center regions are highlighted, while the backgrounds are suppressed. 
To locate the densely-arranged inshore ships, the NAM is constructed to extract and fuse different fine-grained long-range dependencies. It can provide comprehensive global information at pixel-level and target-level, enriching features of varying granularities and enabling the network to distinguish ship's boundaries accurately. 
To accommodate the diverse sizes of ships, the CC-FPN is proposed to expand and integrate feature maps across various levels, thereby bolstering the detection capability for both large and small ships.
Experiments on SSDD, HRSID and LS-SSDD reflect that the mAP of CASS-Det are 0.992, 0.931 and 0.821, respectively. These scores surpass the second highest performing detectors by 0.012, 0.018, and 0.018, marking a significant advancement in performance. 
In the future, we will explore ship detection in high-noise SAR images, further enhancing the practical applicability of our model in real-world scenarios.

\bibliographystyle{IEEEtran}
\bibliography{IEEEabrv,Bibliography}

\begin{thebibliography}{10}
\providecommand{\url}[1]{#1}
\csname url@rmstyle\endcsname
\providecommand{\newblock}{\relax}
\providecommand{\bibinfo}[2]{#2}
\providecommand\BIBentrySTDinterwordspacing{\spaceskip=0pt\relax}
\providecommand\BIBentryALTinterwordstretchfactor{4}
\providecommand\BIBentryALTinterwordspacing{\spaceskip=\fontdimen2\font plus
\BIBentryALTinterwordstretchfactor\fontdimen3\font minus \fontdimen4\font\relax}
\providecommand\BIBforeignlanguage[2]{{%
\expandafter\ifx\csname l@#1\endcsname\relax
\typeout{** WARNING: IEEEtran.bst: No hyphenation pattern has been}%
\typeout{** loaded for the language `#1'. Using the pattern for}%
\typeout{** the default language instead.}%
\else
\language=\csname l@#1\endcsname
\fi
#2}}
\renewcommand\BIBentryALTinterwordstretchfactor{4}

\bibitem{1}
N.~Wang, B.~Li, X.~Wei, Y.~Wang, and H.~Yan, ``Ship detection in spaceborne infrared image based on lightweight cnn and multisource feature cascade decision,'' \emph{IEEE Transactions on Geoscience and Remote Sensing}, vol.~59, no.~5, pp. 4324--4339, 2021.

\bibitem{2}
Y.~Guan, X.~Zhang, S.~Chen, G.~Liu, Y.~Jia, Y.~Zhang, G.~Gao, J.~Zhang, Z.~Li, and C.~Cao, ``Fishing vessel classification in sar images using a novel deep learning model,'' \emph{IEEE Transactions on Geoscience and Remote Sensing}, vol.~61, pp. 1--21, 2023.

\bibitem{3}
X.~Leng, K.~Ji, S.~Zhou, and X.~Xing, ``Ship detection based on complex signal kurtosis in single-channel sar imagery,'' \emph{IEEE Transactions on Geoscience and Remote Sensing}, vol.~57, no.~9, pp. 6447--6461, 2019.

\bibitem{4}
F.~Shen, Y.~Xie, J.~Zhu, X.~Zhu, and H.~Zeng, ``Git: Graph interactive transformer for vehicle re-identification,'' \emph{IEEE Transactions on Image Processing}, vol.~32, pp. 1039--1051, 2023.

\bibitem{5}
Y.~Liu, G.~Yan, F.~Ma, Y.~Zhou, and F.~Zhang, ``Sar ship detection based on explainable evidence learning under intraclass imbalance,'' \emph{IEEE Transactions on Geoscience and Remote Sensing}, vol.~62, pp. 1--15, 2024.

\bibitem{6}
J.~Zhang, M.~Xing, G.-C. Sun, and N.~Li, ``Oriented gaussian function-based box boundary-aware vectors for oriented ship detection in multiresolution sar imagery,'' \emph{IEEE Transactions on Geoscience and Remote Sensing}, vol.~60, pp. 1--15, 2022.

\bibitem{10}
G.~Gao, L.~Liu, L.~Zhao, G.~Shi, and G.~Kuang, ``An adaptive and fast cfar algorithm based on automatic censoring for target detection in high-resolution sar images,'' \emph{IEEE Transactions on Geoscience and Remote Sensing}, vol.~47, no.~6, pp. 1685--1697, 2009.

\bibitem{11}
M.~Yang and C.~Guo, ``Ship detection in sar images based on lognormal $\rho$ -metric,'' \emph{IEEE Geoscience and Remote Sensing Letters}, vol.~15, no.~9, pp. 1372--1376, 2018.

\bibitem{12}
Y.~Deng, H.~Wang, S.~Liu, M.~Sun, and X.~Li, ``Analysis of the ship target detection in high-resolution sar images based on information theory and harris corner detection,'' \emph{Eurasip Journal on Wireless Communications and Networking}, vol. 2018, no.~1, 2018.

\bibitem{13}
M.~Tello, C.~Lopez-Martinez, and J.~Mallorqui, ``A novel algorithm for ship detection in sar imagery based on the wavelet transform,'' \emph{IEEE Geoscience and Remote Sensing Letters}, vol.~2, no.~2, pp. 201--205, 2005.

\bibitem{14}
H.~Shi, Q.~Zhang, M.~Bian, H.~Wang, Z.~Wang, L.~Chen, and J.~Yang, ``A novel ship detection method based on gradient and integral feature for single-polarization synthetic aperture radar imagery,'' \emph{Sensors}, vol.~18, no.~2, 2018.

\bibitem{15}
J.~Ai, Z.~Cao, Y.~Mao, Z.~Wang, F.~Wang, and J.~Jin, ``An improved bilateral cfar ship detection algorithm for sar image in complex environment,'' \emph{Journal of Radars}, vol.~10, no.~4, pp. 499 -- 515, 2021.

\bibitem{19}
Z.~Li, F.~Liu, W.~Yang, S.~Peng, and J.~Zhou, ``A survey of convolutional neural networks: Analysis, applications, and prospects,'' \emph{IEEE Transactions on Neural Networks and Learning Systems}, vol.~33, no.~12, pp. 6999--7019, 2022.

\bibitem{20}
S.~Ren, K.~He, R.~Girshick, and J.~Sun, ``Faster r-cnn: Towards real-time object detection with region proposal networks,'' \emph{IEEE Transactions on Pattern Analysis and Machine Intelligence}, vol.~39, no.~6, pp. 1137--1149, 2017.

\bibitem{21}
K.~Yan, F.~Shen, and Z.~Li, ``Enhancing landslide segmentation with guide attention mechanism and fast fourier transformer,'' in \emph{International Conference on Intelligent Computing}.\hskip 1em plus 0.5em minus 0.4em\relax Springer, 2024, pp. 296--307.

\bibitem{22}
Z.~Cai and N.~Vasconcelos, ``Cascade r-cnn: Delving into high quality object detection,'' in \emph{2018 IEEE/CVF Conference on Computer Vision and Pattern Recognition}, 2018, pp. 6154--6162.

\bibitem{23}
L.~Wei, A.~Dragomir, E.~Dumitru, S.~Christian, R.~Scott, F.~Cheng-Yang, and A.~C. Berg, ``Ssd: Single shot multibox detector,'' \emph{Springer, Cham}, 2016.

\bibitem{24}
J.~Redmon, S.~Divvala, R.~Girshick, and A.~Farhadi, ``You only look once: Unified, real-time object detection,'' in \emph{2016 IEEE Conference on Computer Vision and Pattern Recognition (CVPR)}, 2016, pp. 779--788.

\bibitem{25}
J.~Redmon and A.~Farhadi, ``Yolov3: An incremental improvement,'' \emph{arXiv e-prints}, 2018.

\bibitem{26}
A.~Bochkovskiy, C.~Y. Wang, and H.~Y.~M. Liao, ``Yolov4: Optimal speed and accuracy of object detection,'' 2020.

\bibitem{27}
C.~Y. Wang, A.~Bochkovskiy, and H.~Y.~M. Liao, ``Yolov7: Trainable bag-of-freebies sets new state-of-the-art for real-time object detectors,'' \emph{arXiv e-prints}, 2022.

\bibitem{28}
Z.~Ge, S.~Liu, F.~Wang, Z.~Li, and J.~Sun, ``Yolox: Exceeding yolo series in 2021,'' \emph{arXiv e-prints}, 2021.

\bibitem{29}
T.-Y. Lin, P.~Goyal, R.~Girshick, K.~He, and P.~Dollár, ``Focal loss for dense object detection,'' \emph{IEEE Transactions on Pattern Analysis and Machine Intelligence}, vol.~42, no.~2, pp. 318--327, 2020.

\bibitem{30}
Z.~Tian, C.~Shen, H.~Chen, and T.~He, ``Fcos: Fully convolutional one-stage object detection,'' in \emph{2019 IEEE/CVF International Conference on Computer Vision (ICCV)}, 2019, pp. 9626--9635.

\bibitem{31}
K.~Duan, S.~Bai, L.~Xie, H.~Qi, Q.~Huang, and Q.~Tian, ``Centernet: Keypoint triplets for object detection,'' in \emph{2019 IEEE/CVF International Conference on Computer Vision (ICCV)}, 2019, pp. 6568--6577.

\bibitem{SSE}
Z.~Cui, X.~Wang, N.~Liu, Z.~Cao, and J.~Yang, ``Ship detection in large-scale sar images via spatial shuffle-group enhance attention,'' \emph{IEEE Transactions on Geoscience and Remote Sensing}, vol.~59, no.~1, pp. 379--391, 2021.

\bibitem{Light}
B.~Hu and H.~Miao, ``An improved deep neural network for small-ship detection in sar imagery,'' \emph{IEEE Journal of Selected Topics in Applied Earth Observations and Remote Sensing}, vol.~17, pp. 2596--2609, 2024.

\bibitem{39}
J.~Jiang, X.~Fu, R.~Qin, X.~Wang, and Z.~Ma, ``High-speed lightweight ship detection algorithm based on yolo-v4 for three-channels rgb sar image,'' \emph{Remote Sensing}, vol.~13, no.~10, 2021.

\bibitem{41}
T.~Y. Lin, P.~Dollar, R.~Girshick, K.~He, B.~Hariharan, and S.~Belongie, ``Feature pyramid networks for object detection,'' \emph{IEEE Computer Society}, 2017.

\bibitem{42}
S.~Liu, L.~Qi, H.~Qin, J.~Shi, and J.~Jia, ``Path aggregation network for instance segmentation,'' in \emph{2018 IEEE/CVF Conference on Computer Vision and Pattern Recognition}, 2018, pp. 8759--8768.

\bibitem{43}
J.~Li, C.~Qu, and J.~Shao, ``Ship detection in sar images based on an improved faster r-cnn,'' in \emph{Sar in Big Data Era: Models, Methods \& Applications}, 2017.

\bibitem{44}
S.~Wei, X.~Zeng, Q.~Qu, M.~Wang, H.~Su, and J.~Shi, ``Hrsid: A high-resolution sar images dataset for ship detection and instance segmentation,'' \emph{IEEE Access}, vol.~8, pp. 120\,234--120\,254, 2020.

\bibitem{LS-SSDD}
\BIBentryALTinterwordspacing
T.~Zhang, X.~Zhang, X.~Ke, X.~Zhan, J.~Shi, S.~Wei, D.~Pan, J.~Li, H.~Su, Y.~Zhou, and D.~Kumar, ``Ls-ssdd-v1.0: A deep learning dataset dedicated to small ship detection from large-scale sentinel-1 sar images,'' \emph{Remote Sensing}, vol.~12, no.~18, 2020. [Online]. Available: \url{https://www.mdpi.com/2072-4292/12/18/2997}
\BIBentrySTDinterwordspacing

\bibitem{40}
D.~Pan, X.~Gao, W.~Dai, J.~Fu, Z.~Wang, X.~Sun, and Y.~Wu, ``Srt-net: Scattering region topology network for oriented ship detection in large-scale sar images,'' \emph{IEEE Transactions on Geoscience and Remote Sensing}, vol.~62, pp. 1--18, 2024.

\bibitem{33}
Z.~Cui, Q.~Li, Z.~Cao, and N.~Liu, ``Dense attention pyramid networks for multi-scale ship detection in sar images,'' \emph{IEEE Transactions on Geoscience and Remote Sensing}, vol.~57, no.~11, pp. 8983--8997, 2019.

\bibitem{34}
S.~Woo, J.~Park, J.~Y. Lee, and I.~S. Kweon, ``Cbam: Convolutional block attention module,'' \emph{Springer, Cham}, 2018.

\bibitem{38}
X.~Yang, X.~Zhang, N.~Wang, and X.~Gao, ``A robust one-stage detector for multiscale ship detection with complex background in massive sar images,'' \emph{IEEE Transactions on Geoscience and Remote Sensing}, vol.~60, pp. 1--12, 2022.

\bibitem{BANET}
Q.~Hu, S.~Hu, and S.~Liu, ``Banet: A balance attention network for anchor-free ship detection in sar images,'' \emph{IEEE Transactions on Geoscience and Remote Sensing}, vol.~60, pp. 1--12, 2022.

\bibitem{35}
L.~Zhang, Z.~Chu, and B.~Zou, ``Multi scale ship detection based on attention and weighted fusion model for high resolution sar images,'' in \emph{IGARSS 2022 - 2022 IEEE International Geoscience and Remote Sensing Symposium}, 2022, pp. 631--634.

\bibitem{36}
Q.~Hou, D.~Zhou, and J.~Feng, ``Coordinate attention for efficient mobile network design,'' in \emph{2021 IEEE/CVF Conference on Computer Vision and Pattern Recognition (CVPR)}, 2021, pp. 13\,708--13\,717.

\bibitem{37}
G.~Tang, H.~Zhao, C.~Claramunt, W.~Zhu, S.~Wang, Y.~Wang, and Y.~Ding, ``Ppa-net: Pyramid pooling attention network for multi-scale ship detection in sar images,'' \emph{Remote Sensing}, vol.~15, no.~11, 2023.

\bibitem{shen2024imagpose}
F.~Shen and J.~Tang, ``Imagpose: A unified conditional framework for pose-guided person generation,'' in \emph{The Thirty-eighth Annual Conference on Neural Information Processing Systems}, 2024.

\bibitem{shen2024imagdressing}
F.~Shen, X.~Jiang, X.~He, H.~Ye, C.~Wang, X.~Du, Z.~Li, and J.~Tang, ``Imagdressing-v1: Customizable virtual dressing,'' \emph{arXiv preprint arXiv:2407.12705}, 2024.

\bibitem{shen2024boosting}
F.~Shen, H.~Ye, S.~Liu, J.~Zhang, C.~Wang, X.~Han, and W.~Yang, ``Boosting consistency in story visualization with rich-contextual conditional diffusion models,'' \emph{arXiv preprint arXiv:2407.02482}, 2024.

\bibitem{shen2023advancing}
F.~Shen, H.~Ye, J.~Zhang, C.~Wang, X.~Han, and W.~Yang, ``Advancing pose-guided image synthesis with progressive conditional diffusion models,'' \emph{arXiv preprint arXiv:2310.06313}, 2023.

\bibitem{Non-local}
X.~Wang, R.~Girshick, A.~Gupta, and K.~He, ``Non-local neural networks,'' in \emph{2018 IEEE/CVF Conference on Computer Vision and Pattern Recognition}, 2018, pp. 7794--7803.

\bibitem{GCNet}
Y.~Cao, J.~Xu, S.~Lin, F.~Wei, and H.~Hu, ``Gcnet: Non-local networks meet squeeze-excitation networks and beyond,'' in \emph{2019 IEEE/CVF International Conference on Computer Vision Workshop (ICCVW)}, 2019, pp. 1971--1980.

\bibitem{DAnet}
J.~Fu, J.~Liu, H.~Tian, Y.~Li, Y.~Bao, Z.~Fang, and H.~Lu, ``Dual attention network for scene segmentation,'' 2018.

\bibitem{CCNet}
Z.~Huang, X.~Wang, Y.~Wei, L.~Huang, H.~Shi, W.~Liu, and T.~S. Huang, ``Ccnet: Criss-cross attention for semantic segmentation,'' \emph{IEEE Transactions on Pattern Analysis and Machine Intelligence}, vol.~45, no.~6, pp. 6896--6908, 2023.

\bibitem{SER-Faster}
Z.~Lin, K.~Ji, X.~Leng, and G.~Kuang, ``Squeeze and excitation rank faster r-cnn for ship detection in sar images,'' \emph{IEEE Geoscience and Remote Sensing Letters}, vol.~16, no.~5, pp. 751--755, 2019.

\bibitem{Dynamic}
\BIBentryALTinterwordspacing
H.~Zhang, H.~Chang, B.~Ma, N.~Wang, and X.~Chen, ``Dynamic r-cnn: Towards high quality object detection via dynamic training.''\hskip 1em plus 0.5em minus 0.4em\relax Berlin, Heidelberg: Springer-Verlag, 2020. [Online]. Available: \url{https://doi.org/10.1007/978-3-030-58555-6_16}
\BIBentrySTDinterwordspacing

\bibitem{Double-head}
Y.~Wu, Y.~Chen, L.~Yuan, Z.~Liu, L.~Wang, H.~Li, and Y.~Fu, ``Rethinking classification and localization for object detection,'' in \emph{2020 IEEE/CVF Conference on Computer Vision and Pattern Recognition (CVPR)}, 2020, pp. 10\,183--10\,192.

\bibitem{CR2A-Net}
Y.~Yu, X.~Yang, J.~Li, and X.~Gao, ``A cascade rotated anchor-aided detector for ship detection in remote sensing images,'' \emph{IEEE Transactions on Geoscience and Remote Sensing}, vol.~60, pp. 1--14, 2022.

\bibitem{FCOS}
Z.~Tian, C.~Shen, H.~Chen, and T.~He, ``Fcos: Fully convolutional one-stage object detection,'' in \emph{2019 IEEE/CVF International Conference on Computer Vision (ICCV)}, 2019, pp. 9626--9635.

\bibitem{Free-anchor}
X.~Zhang, F.~Wan, C.~Liu, X.~Ji, and Q.~Ye, ``Learning to match anchors for visual object detection,'' \emph{IEEE Transactions on Pattern Analysis and Machine Intelligence}, vol.~44, no.~6, pp. 3096--3109, 2022.

\bibitem{YOLO-FA}
L.~Zhang, Y.~Liu, W.~Zhao, X.~Wang, G.~Li, and Y.~He, ``Frequency-adaptive learning for sar ship detection in clutter scenes,'' \emph{IEEE Transactions on Geoscience and Remote Sensing}, vol.~61, pp. 1--14, 2023.

\bibitem{Yolov8}
R.~Varghese and S.~M., ``Yolov8: A novel object detection algorithm with enhanced performance and robustness,'' in \emph{2024 International Conference on Advances in Data Engineering and Intelligent Computing Systems (ADICS)}, 2024, pp. 1--6.

\bibitem{Yolox}
\BIBentryALTinterwordspacing
Z.~Ge, S.~Liu, F.~Wang, Z.~Li, and J.~Sun, ``Yolox: Exceeding yolo series in 2021,'' 2021. [Online]. Available: \url{https://arxiv.org/abs/2107.08430}
\BIBentrySTDinterwordspacing

\bibitem{BiFF-FESA}
Y.~Zhou, H.~Liu, F.~Ma, Z.~Pan, and F.~Zhang, ``A sidelobe-aware small ship detection network for synthetic aperture radar imagery,'' \emph{IEEE Transactions on Geoscience and Remote Sensing}, vol.~61, pp. 1--16, 2023.

\bibitem{Yolov7}
C.-Y. Wang, A.~Bochkovskiy, and H.-Y.~M. Liao, ``Yolov7: Trainable bag-of-freebies sets new state-of-the-art for real-time object detectors,'' in \emph{2023 IEEE/CVF Conference on Computer Vision and Pattern Recognition (CVPR)}, 2023, pp. 7464--7475.

\bibitem{ATSD}
C.~Yao, P.~Xie, L.~Zhang, and Y.~Fang, ``Atsd: Anchor-free two-stage ship detection based on feature enhancement in sar images,'' \emph{Remote Sensing}, vol.~14, no.~23, 2022.

\bibitem{DWB-YOLO}
X.~Tang, J.~Zhang, Y.~Xia, and H.~Xiao, ``Dbw-yolo: A high-precision sar ship detection method for complex environments,'' \emph{IEEE Journal of Selected Topics in Applied Earth Observations and Remote Sensing}, vol.~17, pp. 7029--7039, 2024.

\bibitem{45}
H.~Guo, X.~Yang, N.~Wang, B.~Song, and X.~Gao, ``A rotational libra r-cnn method for ship detection,'' \emph{IEEE Transactions on Geoscience and Remote Sensing}, vol.~58, no.~8, pp. 5772--5781, 2020.

\bibitem{DCN}
J.~Dai, H.~Qi, Y.~Xiong, Y.~Li, G.~Zhang, H.~Hu, and Y.~Wei, ``Deformable convolutional networks,'' in \emph{2017 IEEE International Conference on Computer Vision (ICCV)}, 2017, pp. 764--773.

\bibitem{MSIF}
C.~Zhang, C.~Yang, K.~Cheng, N.~Guan, H.~Dong, and B.~Deng, ``Msif: Multisize inference fusion-based false alarm elimination for ship detection in large-scale sar images,'' \emph{IEEE Transactions on Geoscience and Remote Sensing}, vol.~60, pp. 1--11, 2022.

\bibitem{L-YOLO}
G.~Li, X.~Huang, J.~Ai, Z.~Yi, and W.~Xie, ``Lemon‐yolo: an efficient object detection method for lemons in the natural environment,'' \emph{IET Image Processing}, vol.~15, pp. 1998--2009, 2021.

\bibitem{SII-Net}
\BIBentryALTinterwordspacing
N.~Su, J.~He, Y.~Yan, C.~Zhao, and X.~Xing, ``Sii-net: Spatial information integration network for small target detection in sar images,'' \emph{Remote Sensing}, vol.~14, no.~3, 2022. [Online]. Available: \url{https://www.mdpi.com/2072-4292/14/3/442}
\BIBentrySTDinterwordspacing

\bibitem{SAR-Net}
S.~Gao, J.~M. Liu, Y.~H. Miao, and Z.~J. He, ``A high-effective implementation of ship detector for sar images,'' \emph{IEEE Geoscience and Remote Sensing Letters}, vol.~19, pp. 1--5, 2022.

\end{thebibliography}

\end{document}